# Clustering Algorithm for Gujarati Language

**Miral Patel[1], Prem Balani[2]**
[1]Research Scholar, [2]Assistant Professor
[1, 2]GCET, Vallabh Vidyanagar, Anand, Gujarat, India
[1]milipatel8@gmail.com [2]prembalani@gcet.ac.in

*Abstract*— Natural language processing area is still under research. But now a day it is on platform for worldwide researchers. Natural language processing includes analyzing the language based on its structure and then tagging of each word appropriately with its grammar base. Here we have 50,000 tagged words set and we try to cluster those Gujarati words based on proposed algorithm, we have defined our own algorithm for processing. Many clustering techniques are available Ex. Single linkage , complete, linkage ,average linkage, Hear no of clusters to be formed are not known, so it's all depends on the type of data set provided .Clustering is preprocess for stemming . Stemming is the process where root is extracted from its word. Ex. cats= cat+S, meaning. Cat: Noun and plural form.

## I. INTRODUCTION

In NLP :*Natural language processing* is a field of computer science, artificial intelligence, and linguistics concerned with the interactions between computers and human (natural) languages. As such, NLP is related to the area of human–computer interaction. Many challenges in NLhkP involve natural language understanding -- that is, enabling computers to derive meaning from human or natural language input. [I]

*Linguistics* is the scientific study of human language. Linguistics can be broadly broken into three categories or subfields of study: language form, language meaning, and language in context.

*Stemming* is the process where we extract root words from grammar based words for example *grAhako= graham + o*, *mAhiwinI= mAhiwi+nI*. Here we have shown examples, both are Noun. we have many categories in grammar like Adverb, Adjective , Pronoun, Conjuction etc....For clustering preprocessing step is *POS*:. Part of speech tagging POS process will tag each sentence with its grammatical identifier for Example: *mAhiwinI* 'NN - here 'NN shows NOUN category. Similarly all input test corpus is tagged with its label.

There are two approaches available Supervised and unsupervised. Many clustering techniques are available like single linkage, average linkage, and complete linkage. The similarity between two groups is defined as the maximum similarity between any member of one group and any member of the other. Groups only need to be similar in a single pair of members in order to be merged [3]

The similarity of two clusters is calculated as the minimum similarity between any member of one cluster and any member of the other. Like single linkage, the probability of an element merging with a cluster is determined by a single member of the cluster. However, in this case the least similar member is considered, instead of the most [3].

The similarity between two groups of points is defined by the mean similarity between points in one cluster and those of the other. In contrast to a single linkage each element needs to be relatively similar to all members of the other cluster, rather than to just one. Average linkage clusters tend to be relatively round or ellipsoid. [3]

For clustering process we have used a supervised approach as we had fixed no of category for clustering. we have worked for categories are listed in table 1.

| Category | 50,529 |
|---|---|
| NN | Noun |
| JJ | Adjective |
| PRP | Preposition |
| PSP | Post position |
| CC | Conjuction |
| VM | Verb Main |
| VAUX | Verb Auxiliary |
| NNC | Special symbol |

Table. 1: List of Tags

## II. PROPOSED ALGORITHAMS FOR CLUSTERING IN GUAJARATI

**Step 1**: Create an object for File input strem ,Datainput stream and bufferreader classes "fstream","in","br" respectively . Open the file "pos.txt" in read mode .

**Step 2**: Create an object of file writer and buffer reader classes "Nounstream" "noun" respectively

*FileWriter nounstream = new FileWriter("c:\\noun.txt");*

*BufferedWriter noun = new BufferedWriter(nounstream);*

**Step 3**: Repeat Step 2 for all categories and create objects of file writer and Bufferrederclasses.jjstream,jj, Vmstream, vm, vauxstream, vaux, ccstream, cc, nnstream, nn, prpstream, prp, pspstream, psp. Nncstream, nnc & open the respected files in write mode .exmaple jj.txt

**Step 4**: read "Pos.txt" file . Br.ReadLine;

**Step 5**: Split lines to the word format and store in to String array M.

**Step 6**: Count length of array

6.1 Read word.

6.2 check its postposition

6.3 Define boolean flags &Set flags accordingly

```
        boolean returnnn ;
        boolean returnjj ;
        boolean returnvm ;
        boolean returnvaux ;
        boolean returncc ;
        boolean returnpsp;
        boolean returnnnc ;
        boolean returnprp;
   check flags
        if (returnnn ==“NN”)
              write to “NN .txt”
        else if (returnjj==“JJ”)
```

```
write to "jj.txt"
        else if (returnvm=="vm")
                write to  "vm.txt"
        else   if (returnvaux =="Vaux")
                write to "vaux.txt"
        else if (returncc=="CC")
                write to "cc.txt"
        else if (returnpsp=="PSP")
                write to  "psp.txt"
        else if (returnpsp=="NNC")
                write to  "nnc.txt"
        else if (returnpsp=="PRP")
                write to  "prp.txt"
    6.4  Close all input and output stream ex: in.close();
    // try block over
}
    Step 7: catch exception if any ..//depends on type of io error
} // main fucntion over
// class closed.
```

As shown in algorithm We had an input file of 50,000 Guajarati words to the system. POS process is applied as first step to tag the words according to its grammar base of gujarati like noun , Adjective adverb Once we have resultant file of tagged word we use that file as input for clustering process and as a result it provides different text file based on different category.

## III. RESULTS AND EVALUATION

The results of this algorithm are shown in table. 2This method will work perfect for clustering of tagged words. if the corpus is directly used with untagged words this method is not suitable. For untagged words any distance matrix based statistical method can be used. Using supervised method we found almost all tagged words are correctly classified in no of clusters. Here no of clusters will be no of category of grammar we have taken. We have used standard tag library for implementation purpose. One may use different tag set based on different language and different library .Corpus we used is of 50,529 words .one may try to increase corpus and check results.

| Clustering words | 50,529 |
|---|---|
| nn | 13780 |
| jj | 2862 |
| prp | 5508 |
| psp | 3133 |
| cc | 3182 |
| vm | 6747 |
| vaux | 3304 |
| Nnc | 624 |

Table. 2: Results of clustering

As shown in table 2 clustering process will create a different text file for different category of tag. Each file contains no tag as shown in table we have found this algorithm works successfully with tagging words and so accuracy is 98% if some of the words are not tagged properly then there are prey good chances of error in clustering process otherwise on correctly tagged words we found accuracy more than 98%.

Now this file can be used as input file for stemming process. So we can conclude here that clustering is a necessary and an intermediate step in stemming.

As future work one may try to modify the same algorithm for untagged word and can evaluate results. One may try to cluster using an unsupervised approach and can check the efficiency of proposed algorithms. Following are text file snapshot taken for result evaluation. All implementations are done in JAVA and for Gujarati language WX notation used.

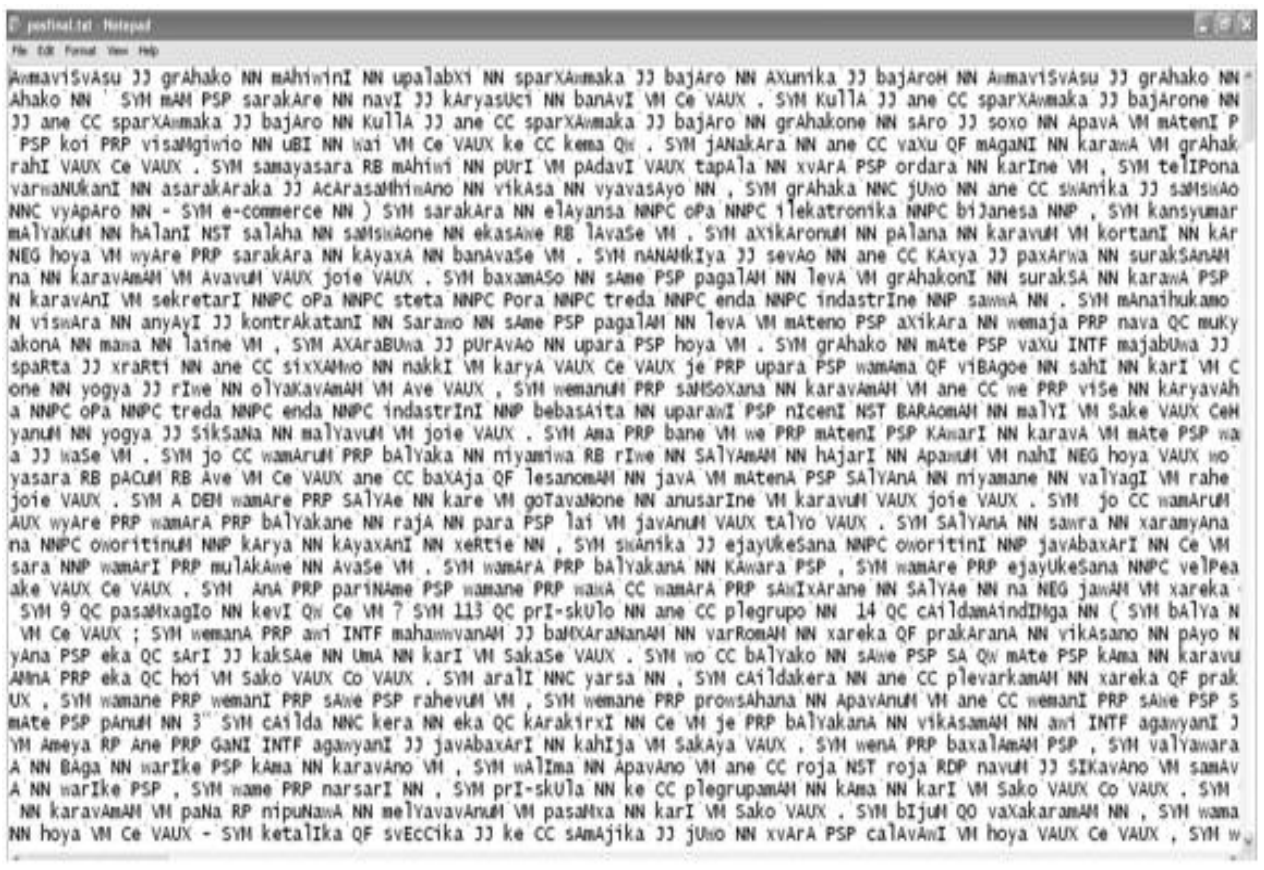

Fig. 1: Input files For Clustering

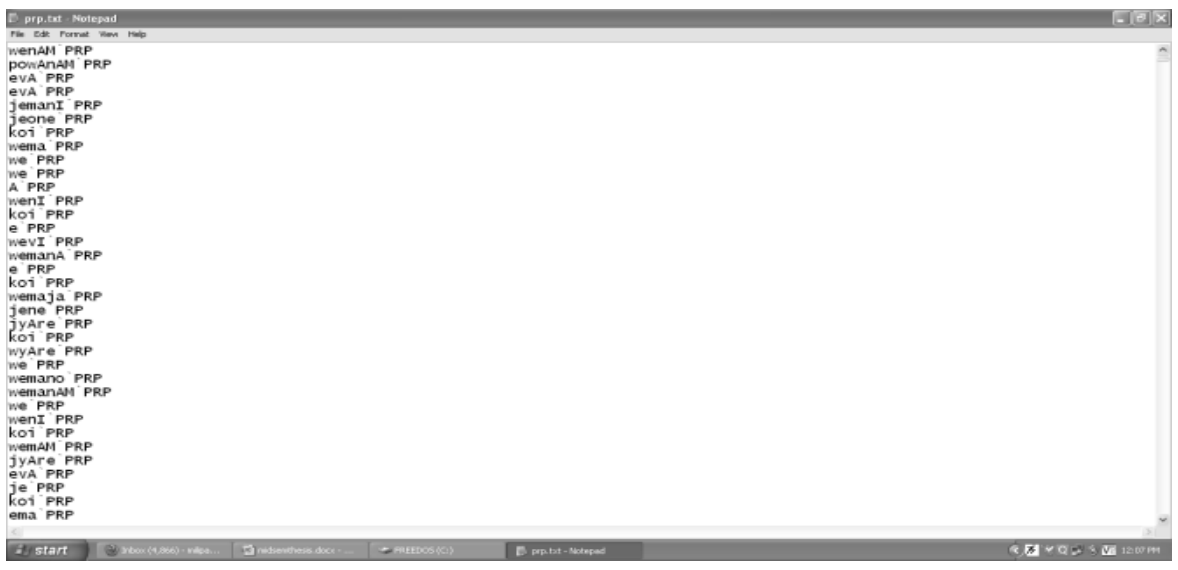

Fig. 2(a): Output files: Category shown in table. 2

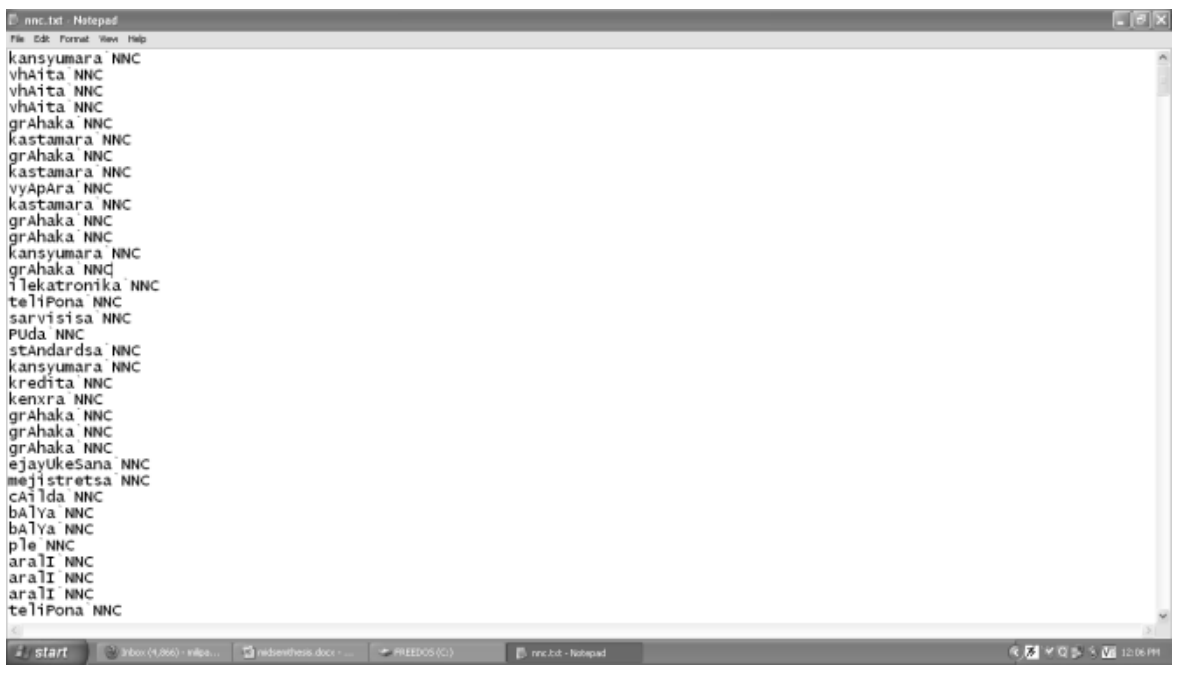

(b)

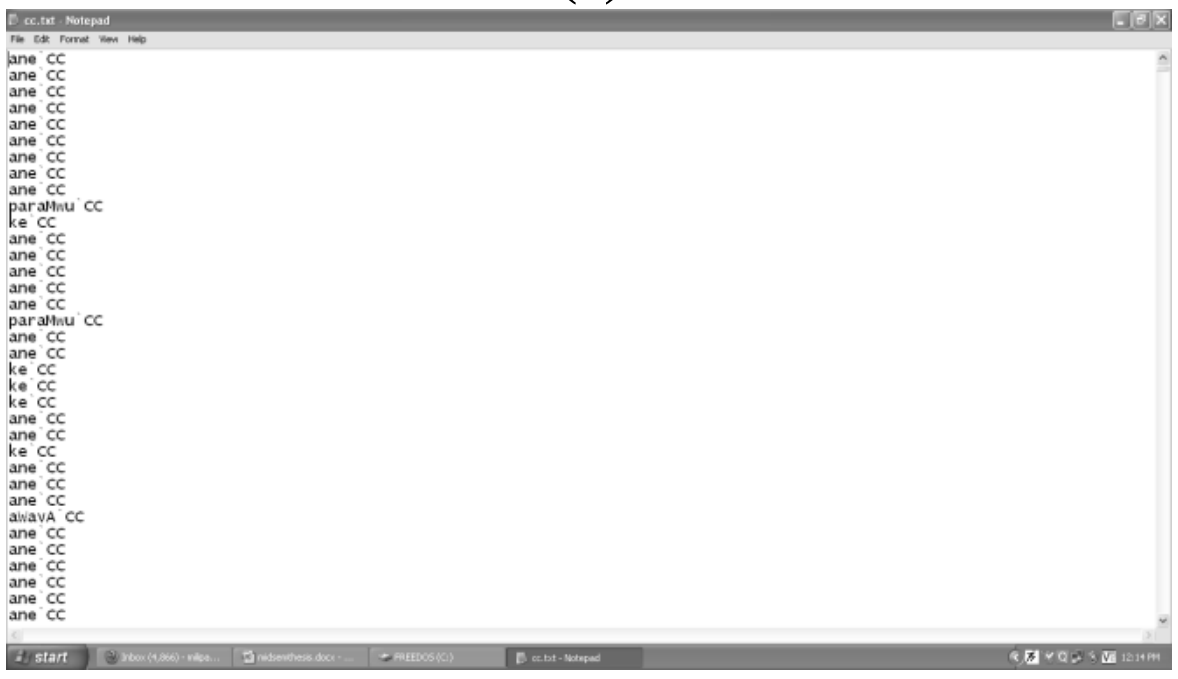

(c)

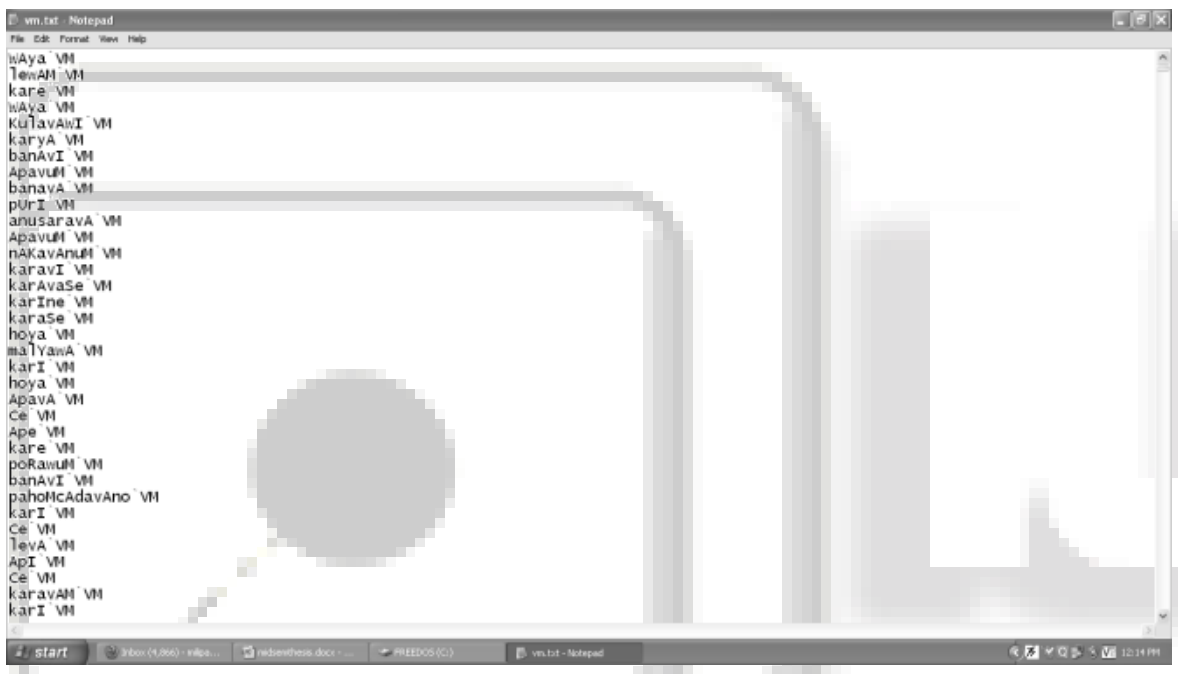

(d)

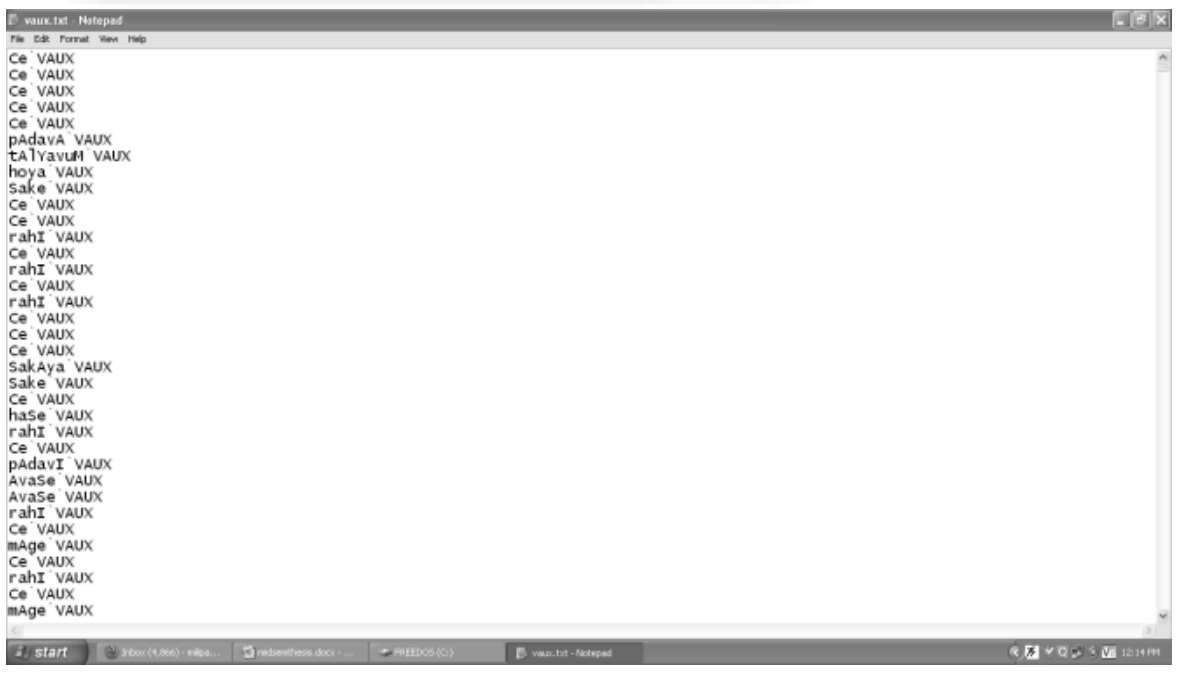

(e)

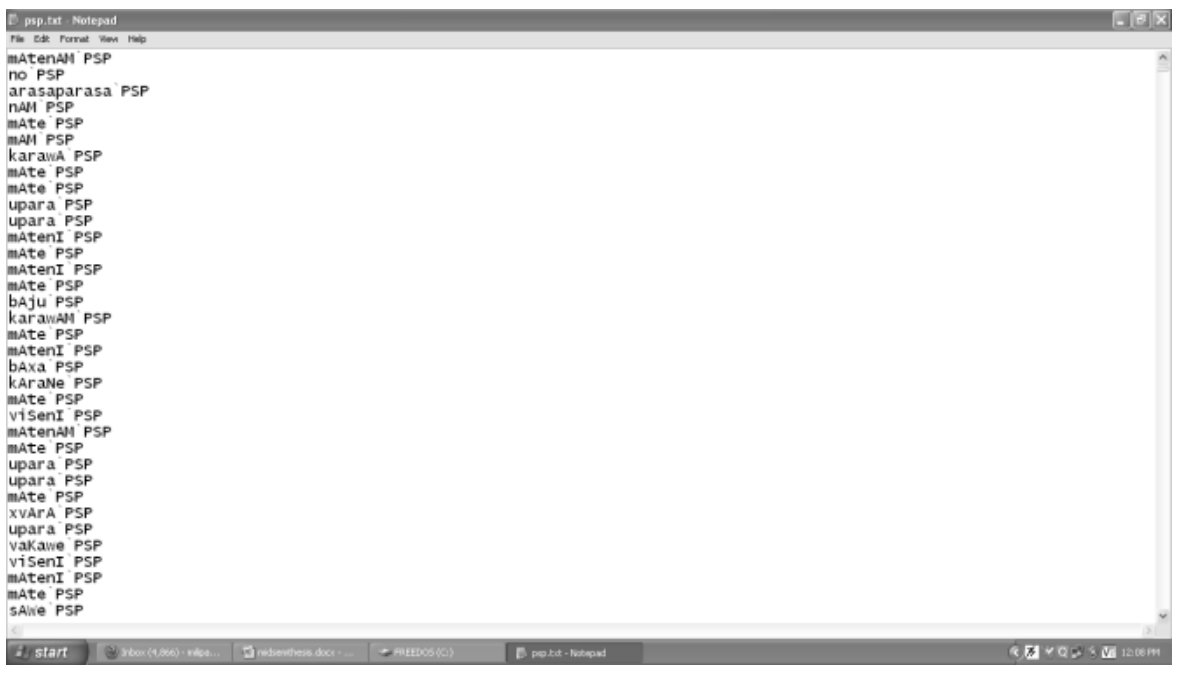

(f)

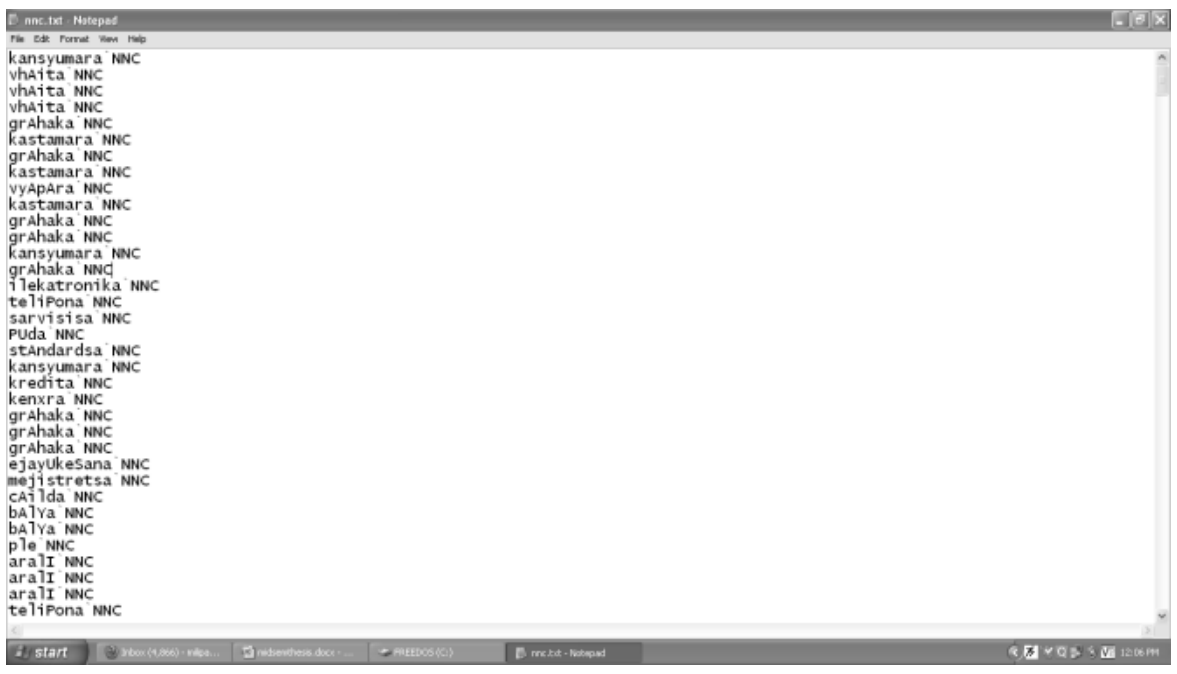

(g)

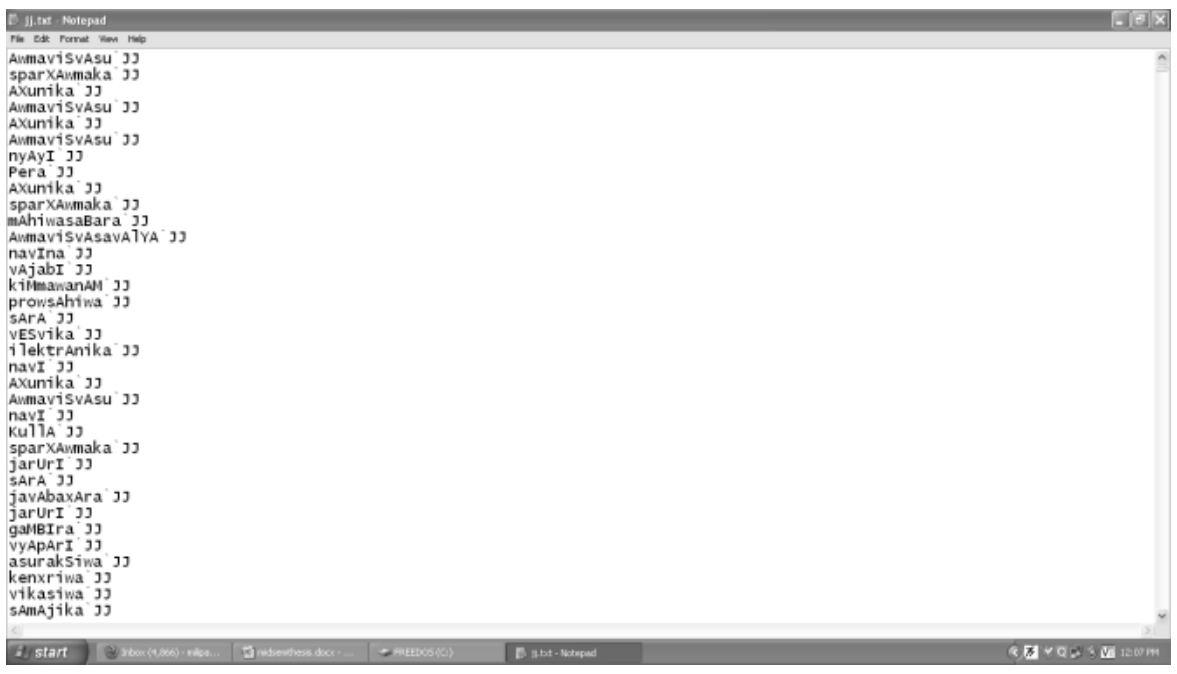

(h)

Fig. 2 (a) – (h) Output files: Category shown in table (1.2)

## IV. CONCLUSION AND FUTURE WORK

The results are pretty clear after clustering of tagging words, but one may work on different language and different tag sets for clustering. Here the conclusion is clustering process give more than 98% accuracy if it is done with tagged words and clustering on tagged word is an intermediate step for to reach stemming process.

## REFERENCES

[1]. Ramnathan, D Rao, *"A Lightweight Stemmer for Hindi"*, *In* Proceedings of Workshop on Computational Linguistics for South Asian Languages, 10th Conference of the European Chappter of Association of Computational Linguistics. pp 42-48. 2003.

[2]. Juhi Ameta, Nisheeth Joshi, Iti Mathur,"*A Lightweight Stemmer for Gujarati*"
Department of Computer Science, Apaji Institute, Banasthali University, Rajasthan, India,

[3]. Majumder, Prasenjit, MandarMitra, Swapan K. Parui, GobindaKole, PabitraMitra, and KalyankumarDatta. 2007. YASS: *Yet another suffix stripper*. *Association for Computing Machinery Transactions on Information Systems*, 25(4):pp 18-38.

[4]. Mohd. Shahid Husain, August 2012. *An Unsupervised approach to develop stemmer International Journal on Natural Language Computing* (IJNLC) Vol. 1, No.2, August 2012Department of Information Technology, Integral University, Lucknow.

[5]. Rashmi S, Kratika Singh, Ajay Dhamelia "*Stemming and Morphological Analysis*"

[6]. Zuckermann, Ghil'ad 2003, *Language Contact and Lexical Enrichment in Israeli Hebrew*, *Houndmills*: Palgrave Macmillan. ISBN 1-4039-1723-X. pp 65–66.

[7]. Lovins, Julie Beth (1968). *"Development of a Stemming Algorithm". Mechanical Translation and Computational Linguistics 11*: 22–31.
[8]. Powers, David M W (2007/2011). *"Evaluation: From Precision, Recall and F-Factor to ROC, Informedness, Markedness & Correlation". Journal of Machine Learning Technologies 2* (1): 37–63
[9]. http://en.wikipedia.org/wiki/Root_%28linguistics%29
[10]. http://en.wikipedia.org/wiki/Stemming
[11]. http://en.wikipedia.org/wiki/Linguistics

WX NOTATION`

| | | | | | | | | | | | | | | |
|---|---|---|---|---|---|---|---|---|---|---|---|---|---|---|
| a | aA | ai | aI | au | aU | aeV | ae | aE | aEY | aoV | ao | aO | aOY | |
| अ | आ | इ | ई | उ | ऊ | ऎ | ए | ऐ | ऍ | ऒ | ओ | औ | ऑ | |
| aM | aH | az | aq | aQ | | | | | | | | | | |
| अं | अः | अँ ँ | ऋ | ॠ | | | | | | | | | | |
| A | i | I | u | U | eV | e | E | EY | oV | o | O | OY | M | H |
| ा | ि | ी | ु | ृ | ॆ | े | ै | ॅ | ॊ | ो | ौ | ॉ | ं | ः |
| k | K | g | G | f | | | | | | | | | | |
| क | ख | ग | घ | ङ | | | | | | | | | | |
| c | C | j | J | F | | | | | | | | | | |
| च | छ | ज | झ | ञ | | | | | | | | | | |